\documentclass[sensors,article,moreauthors,pdftex]{Definitions/mdpi} 
\renewcommand{\linenumbers}{}
\usepackage{glossaries}
\usepackage{xcolor}
\usepackage{gensymb}
\usepackage{soul}

\usepackage{subcaption}
\usepackage{pgf}
\usepackage[utf8]{inputenc}\DeclareUnicodeCharacter{2212}{-}

\newacronym{CABE}{CABE}{Cloud based Acoustic Beamforming Emulator}
\newacronym{gcd}{GCD}{Greatest Common Divisor}
\newacronym{FOV}{FOV}{Field of View}

\firstpage{1} 
\pubvolume{xx}
\issuenum{1}
\articlenumber{5}
\pubyear{2021}
\copyrightyear{2021}
\history{Received: date; Accepted: date; Published: date}

\Title{XCycles Backprojection Acoustic Super-Resolution}

\Author{ {Feras Almasri} 
 $^{1,2,}$*\orcidA{}, Jurgen Vandendriessche $^{1}$\orcidB{}, Laurent Segers $^{1}$\orcidC{}, Bruno da Silva $^{1,3,}$*\orcidD{}, An Braeken $^{1}$\orcidE{}, Kris~Steenhaut $^{1,3}$\orcidH{}, Abdellah Touhafi $^{1,3}$\orcidF{} and Olivier Debeir $^{2}$\orcidG{}}

\AuthorNames{Feras Almasri, Jurgen Vandendriessche, Laurent Segers, Bruno da Silva, An Braeken, Kris Steenhaut, Abdellah Touhafi and Olivier Debeir}
\address{%
$^{1}$ \quad Department of Engineering Sciences and Technology (INDI), Vrije Universiteit Brussel,  {1050~Brussels},~{Belgium}
; Jurgen.Vandendriessche@vub.be (J.V.); laurent.segers@vub.be (L.S.); abraeken@gmail.com~(A.B.); ksteenha@etrovub.be (K.S.); abdellah.touhafi@vub.ac.be (A.T.)\\
$^{2}$ \quad Laboratory of Image Synthesis and Analysis (LISA), Université Libre de Bruxelles,  {1050 Brussels}
, Belgium; odebeir@ulb.ac.be\\
$^{3}$ \quad Department of Electronics and Informatics (ETRO), VUB,  {1050 Brussels}, Belgium
}
\corres{Correspondence: falmasri@ulb.ac.be (F.A.); Bruno.da.silva@vub.be (B.d.S.)}

\abstract{The computer vision community has paid much attention to the development of visible image super-resolution (SR) using deep neural networks (DNN) and has achieved impressive results. The advancement of non-visible light sensors, such as acoustic imaging sensors, has attracted a lot of attention, as they allow people to visualize the intensity of sound waves beyond the visible spectrum. However, because of the limitations imposed on acquiring acoustic data, new methods for improving the resolution of the acoustic images are necessary. At this time, there is no acoustic imaging dataset designed for the SR problem. 
This work proposes a novel BackProjection model architecture for the acoustic image super-resolution problem, together with Acoustic Map Imaging VUB-ULB Dataset (AMIVU). The dataset provides large simulated and real captured images at different resolutions. The proposed XCycles BackProjection model (XCBP), in contrast to the feedforward model approach, fully uses the iterative correction procedure in each cycle to reconstruct the residual error correction for the encoded features in both low- and high-resolution space. The proposed approach is evaluated on the dataset and showed high outperformance compared to the classical interpolation operators and to the recent feedforward state-of-the-art models. It also contributes to a drastically reduced sub-sampling error produced during the data acquisition.}


\begin{document}

\section{Introduction}

Single-image upscaling, known as Single-Image Super-Resolution (SISR), is a classical computer vision problem, used to increase the spatial resolution of digital images. The process aims to recover fine detail High-Resolution (HR) image from a single coarse Low-Resolution (LR) image. It is an inherently ill-posed inverse problem, as multiple downsampled HR images could correspond to a single LR image. Reciprocally, upsampling an LR image with a 2$\times$ scale factor requires mapping one input value into four values in the high-resolution image, which is usually intractable. It has a wide range of applications, such as digital display technologies, remote sensing, medical imaging care, and security and data content processing. Classical upscaling methods, based on interpolation operators, have been used for this problem for decades, and it remains an active topic of research in image processing. Despite their achieved progress, the upscaled images are still lacking fine details in texture-rich areas. Recently, the development of deep learning methods has witnessed remarkable progress achieving performance on various benchmarks \cite{dong2015image, kim2016accurate} both quantitatively and qualitatively with enhanced details in texture-rich areas.

The computer vision community has paid more attention to the development of visible sensor images, and over the last decade, non-visible light sensors such as infrared, acoustic, and depth imaging sensors were only used in very specific applications.~The images produced by those sensors lack the consideration of the potential benefits of the non-visible spectrum due to their low spatial resolution, the high cost incurred as their price increases dramatically with the increase of their resolution \cite{almasri2018multimodal}, and the lack of publicly available datasets. At this time, there is no acoustic imaging dataset designed for the SR problem. With the need to develop a vision-based system that benefits from the non-visible spectrum, acoustic sensors have recently received much attention as they allow visualization of the intensity of sound waves. The sound intensity in an acoustic heat map format can be graphically represented in order to facilitate the identification and localization of sound~sources.

In contrast to visible or infrared cameras, there is not a single sensor for acoustic imaging, but rather an array of sensors. As a result, the image resolution in acoustic cameras is directly related to their computational demand, requiring hardware accelerators such as Graphical Processor Units (GPUs)~\mbox{\cite{frechette2020low}} or Field-Programmable Gate Arrays (FPGAs)~\mbox{\cite{dasilva2018multimode,Vandendriessche2021M3AC}}. Consequently, available acoustic cameras offer a relatively low resolution, ranging from $40 \times 40$ to $320 \times 240$ pixels per image and at a relatively low frames-per-second rate~\mbox{\cite{zimmermann2010FPGA, izquierdo2016design}}. Acoustic imaging presents a high computational cost and is therefore often prohibitive for embedded systems without hardware accelerators.~Moreover, it also suffers from a subsampling error in the phase delay estimation.~As a result, there is a significant degradation in the quality of the output acoustic image~\mbox{\cite{grondin2019svd}}, which manifests in artifacts that directly affect the sound source localization~\mbox{\cite{zotkin2004accelerated}}. Due to the limitation in acoustic data acquisition, methods to enhance the precision of a measurement with respect to spatial resolution and to reduce artifacts becomes more important.

Learning-based SISR methods rely on high- and low-resolution image pairs generated artificially. The conventional approach is generally by downsampling the images using a bicubic interpolation operator and adding both noise and blur to generate corresponding low-resolution images. However, such image pairs do not exist in real-world problems, and the artificially generated images have a considerable number of artifacts from smoothing, removing sensor noise, and other natural image characteristics. This poses a problem as models trained on these images cannot be generalized to another unknown source of image degradation or natural image characteristics. There are a few contributions where the image pairs were obtained from different cameras in the visible imaging, but none in the acoustic imaging.

Based on these facts, the main contributions of the proposed work are threefold:

\begin{itemize}
\item {A novel backprojection model architecture was proposed to improve the resolution of the acoustic images. The proposed XCycles BackProjection model (XCBP), in contrast to the feedforward model approach, fully uses the iterative correction procedure. It takes low- and high-resolution encoded features together to extract the necessary residual error correction for the encoded features in each cycle.}

\item {The acoustic map imaging  {dataset} (\url{https://doi.org/10.5281/zenodo.4543786}),
 provided simulated and real captured images with multiple scales factor ($\times$2, $\times$4, $\times$8). Although these images shared similar technical characteristics, they lacked artificial artifacts caused by the conventional downsampling strategy. Instead, they had more natural artifacts simulating the real-world problem. The dataset consisted of low- and high-resolution images with double-precision fractional delays and sub-sampling phase delay error.~To the best of the authors' knowledge, this is the first work to provide such a large dataset with its specific characteristics for the SISR problem and the sub-sampling phase delay error problem;}
\item {The proposed benchmark and the developed method outperformed the classical interpolation operators and the recent feedforward state-of-the-art models and drastically reduced the sub-sampling phase delay error estimation.}
\item {\textls[-5]{The proposed model contributed to the Thermal Image Super-Resolution Challenge---}PBVS 2021 \cite{rivadeneira2021thermal} and won the first place with superior performance in the second evaluation when the LR image and HR image are captured with different cameras.} 
\end{itemize}

\section{Related Work}

Plentiful methods have been proposed for image SR in the computer vision community. The original work of introducing Deep Learning based methods for the SR problem by Dong et al \cite{dong2015image} opened new horizons in this SR problem domain. Their proposed model SRCNN achieved superior performance against all previous works. The architecture formulation of the SRCNN aims to learn hierarchical sequence of encoded features and upsample them to reconstruct the final SR image, where the entire learning process can be seen as an end-to-end feedforward manner. In this work, we focus on works related to convolutional neural network (CNN) architecture formulations proposed for residual error correction approach in contrast to the feedforward manner.

The SRCNN aims to lean an end-to-end mapping function between the Bicubic upsampled image and its correspondent high-resolution image, where the last reconstruction layer serves as an encoder from the features space to the super-resolved image space. To speed up the procedure and to reduce the problem complexity, which is proportional to the input image size, Dong et al. \cite{dong2016accelerating} and Shi et al. \cite{shi2016real} proposed faster and learnable upsampling models. Instead of using an interpolated image as an input, they speed up the procedure by reducing the complexity of the encoded features while training upsampling modules at the very end of the network.

The HR image can be seen as components of low-frequency (LF) features (coarse image) and high-frequency (HF) features (residual fine detail image). The results of the classical interpolation operators and the previous deep learning based models have high peak signal-to-noise ratios (PSNR), but they are lacking HF features \cite{ledig2017photo}. As the super-resolved image contains the LF features, Kim et al. \cite{kim2016accurate} proposed the VDSR model that predicts the residual HF features and adds them to the coarse super-resolved image. Their proposed model showed superior performance compared to the previous approaches. Introducing residual networks and skip-connections with residual features correction exhibit improved performance in the SR problem \cite{kim2016accurate, kim2016deeply, ledig2017photo, lim2017enhanced, zhang2018image}, which also allows the end-to-end feedforward to have deeper networks.

In contrast to this simple feedforward approach, Irani et al. \cite{irani1991improving} proposed a model that reconstructs the back projected error in the LR space and adds the residual error to the super-resolved image. Influenced by this work, Haris et al. \cite{haris2018deep} proposed an iterative error correction feedback mechanism to guide the reconstruction of the final SR output. Their approach is a model of iterative up- and downsampling blocks that takes all previously upscaled features and fuses them to reconstruct the SR image. Inspired by \cite{irani1991improving, haris2018deep}, the VCycles Backprojection Upscaling Network (VCBP) ~\cite{rivadeneira2020thermal} was first introduced in the Perception Beyond the Visible Spectrum (PBVS2020) challenge. The model is designed in iterative modules with shared parameters to produce a light SR model dedicated for thermal applications, which limits its performance. In VCBP, the iterative error correction happens in the low-resolution space, and in each iteration, the reconstructed residual features are added to the HR encoded feature space. In VCBPv2 \cite{wei2020aim} the parameters are not shared within the modules, and the iterative error correction happens in both low- and high-resolution space. It follows the design of Haris et al. \cite{haris2018deep} of using iterative up and downsampling layers to process the features in the Inner Loop. Although this technique increases the efficiency of the model, it restricts the depth from extracting important residual features from the encoded features space.

Iterative error correction mechanism in the features space is very important for the HF features reconstruction in the SR problem. If the model pays more attention to this correction procedure in the upsmapled HR and the LR features space, it might be possible to obtain improvements in the residual HF detail in the super-resolved image. This paper proposes a fully iterative backprojection error mechanism network to reconstruct the residual error correction for the encoded features in both low- and high-resolution spaces.

\section{Acoustic Beamforming}
\label{sec:acousticBeamforming}

Acoustic cameras acquire their input signal from arrays of microphones placed in well defined patterns for high spatial resolution.
Microphone arrays come in different shapes and sizes, offering different possibilities for locating a neighbouring sound source.
The process to locate a sound source with a microphone array is referred as the beamforming method.
Beamforming methods comprise several families of algorithms, including the Delay-and-Sum (DaS) beamformers \cite{tashev2005new,soundcompass,Taghizadeh}, the Generalized Sidelobe Cancellation (GSC) beamformers \cite{herbordt2001computationally,lepauloux2010computationally,rombouts2008generalized}, beamformers based on the MUltiple SIgnal Classification (MUSIC) algorithm \cite{gao2018modified,birnie2019sound} and beamformers based on the Estimation of Signal Parameters via Rotational Invariance Technique (ESPRIT) algorithm \cite{jo2018direction,chen2018direction}.


\subsection{Delay and Sum Beamforming}
\label{sec:DaS}

\begin{figure}[ht]
    \centering
    \includegraphics[width=1.0\textwidth]{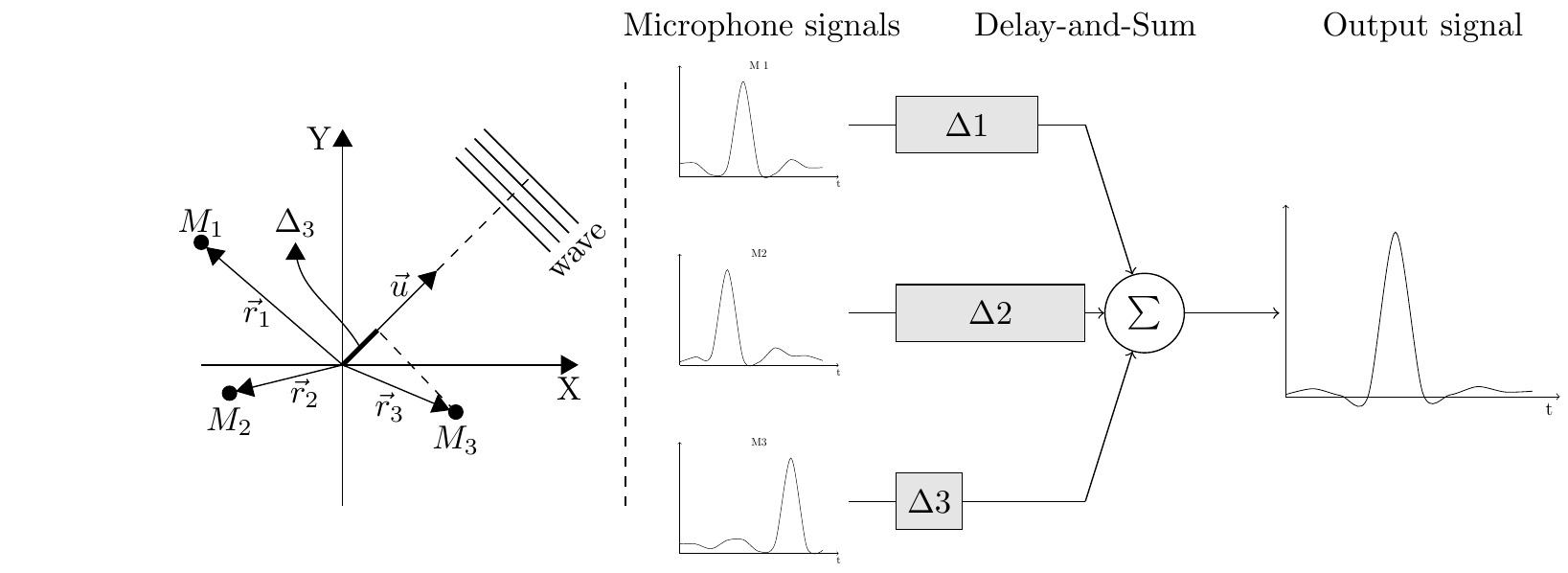}
    \caption{Principle of acoustic beamforming based on the Delay and Sum method.}
    \label{fig:delay_and_sum}
\end{figure}

The well known DaS is the most popular beamformer and is the one selected to generate the acoustic images in this work.
The beamformer steers the microphone array in a particular direction by delaying samples from the different microphones.
The acoustic signals originating from the steering direction are amplified, while acoustic signals coming from other directiosn are suppressed.
The principle of DaS beamforming can be seen in Figure~\mbox{\ref{fig:delay_and_sum}} as:

\begin{equation}
o(\vec{u},t)=\sum\limits_{m=0}^{M-1}s_m(t-\Delta_m(\vec{u}))
\label{eq:steering_all_directions}
\end{equation}

Here, $o(\vec{u},t)$ is the output of the beamformer for a microphone array of $M$ microphones and $s_m(t-\Delta_m)$ is the sample from microphone $m$ delayed by a time $\Delta_m$.
The time delay $\Delta_m(\vec{u})$ in a given steering direction is obtained by computing the dot product between the vector $\vec{r}_m$, describing the location of microphone $m$ in the array, and the unitary steering vector $\vec{u}$. The delay factor is normalized by the speed of sound ($c$) in air. 

\begin{equation}
\Delta_m(\vec{u})=\frac{\vec{r}_m\cdot \vec{u}}{c} 
\label{eq:delay_m}
\end{equation}

The principle described is valid in the continuous time domain. Therefore, the principle of DaS is extended with the sampling frequency $F_s$ so that a form of index $\Delta'_m(\vec{u})$ is obtained:

\begin{equation}
\Delta'_m(\vec{u})=F_s\cdot\frac{\vec{r}_m\cdot \vec{u}}{c}
\label{eq:discrete_delta_m}
\end{equation}

This $\Delta'_m(\vec{u})$ value can be rounded to the nearest integer $\Delta'_{m,round}(\vec{u})$ to facilitate array indexing:

\begin{equation}
\Delta'_{m,round}(\vec{u})=round\left(F_s\cdot\frac{\vec{r}_m\cdot \vec{u}}{c}\right)
\label{eq:discrete_delta_m_rounded}
\end{equation}

For sufficiently high sampling frequencies $F_s$ at the DaS stage, $\Delta'_{m,round}(\vec{u})$ offers a fine grained indexing nearing $\Delta'_m$ so that: 

\begin{equation}
\Delta'_{m,round}(\vec{u})\approx\Delta'_m(\vec{u})
\label{eq:discrete_delay_m_high_fs}
\end{equation}

Based on these equations, the output values $o[\vec{u},i]_{rounded}$ of the DaS method in the time domain with current reference sample index $i$ yields:

\begin{equation}
o[\vec{u},i]_{rounded}=\sum\limits_{m=0}^{M-1}s_m[i-\Delta'_{m,round}(\vec{u})].
\label{eq:delaysumoutput}
\end{equation}

Equation \ref{eq:delaysumoutput} can be transformed into the z-domain by applying the z-domain delay identity so that:

\begin{equation}
O(\vec{u},S,z)=\sum\limits_{m=0}^{M-1}S_m(z)\cdot z^{-\Delta'_{m,round}(\vec{u})}.
\label{eq:z_steering_all_directions}
\end{equation}

From Equation \ref{eq:z_steering_all_directions} one can compute the average Steered Response Power (SRP) $P(\vec{u},S,z)$ over $L$ samples for each of the steering vectors with: 

\begin{equation}
P(\vec{u},S,z) = \frac{1}{L}\sum\limits_{k=0}^{L-1}\left|O(\vec{u},S,z)[k]\right|^2.
\label{eq:z_power}
\end{equation}

By computing the SRP values for each of the steering vectors, a matrix of SRP values can be generated. When all steering vectors have the same elevation, the matrix will be one dimensional and a polar plot can be used for visualisation and finding the origin of the sound source. On the other hand, when the steering vectors have a changing elevation, the matrix will be two dimensional. When this matrix is normalised and, optionally, a colormap is applied to it, the acoustic heatmap is obtained. 

\begin{figure}
    \centering
    \includegraphics[width=0.9\textwidth]{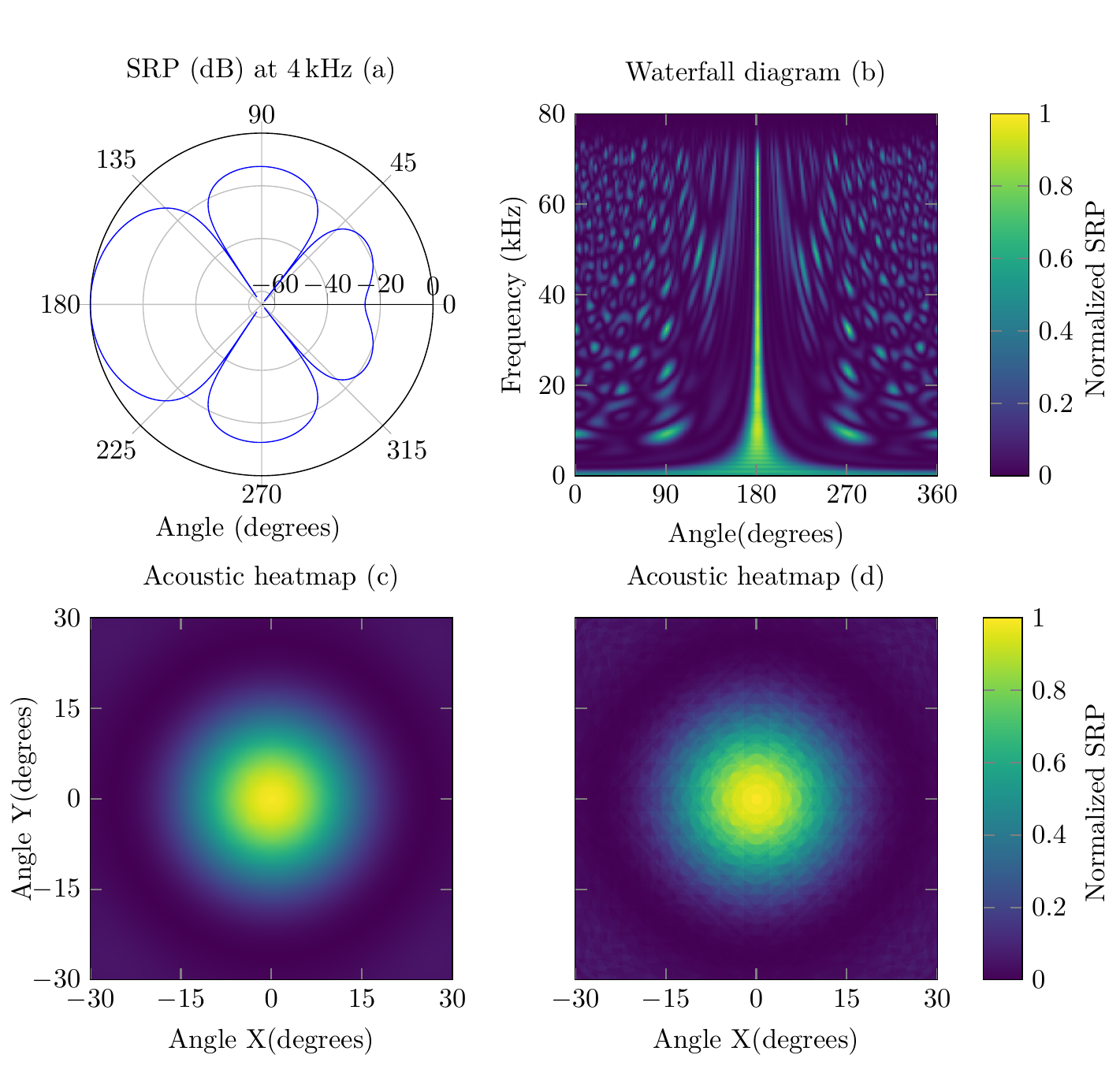} 
    \caption{Examples of an SRP obtained with the microphone array described in Section~\ref{sec:dataset}. A traditional 2-dimensional SRP of a single frequency pointing to an angle of $180^{\circ}$ can be obtained (a). By combining multiple frequencies into one waterfall diagram, one can visualize the frequency response of a given microphone array (b). An acoustic heatmap of a 3D situation (c and d) can also be obtained where the yellow color depicts the highest probability of finding a sound source.  The first heatmap (c) is obtained with fractional delays in double precision format, whereas the last heatmap (d) is obtained without fractional delays.}
\label{fig:steering_examples}
\end{figure}

Examples of the SRP are depicted in Figure \ref{fig:steering_examples}. In (a), a regular beamforming in 2 dimensions is computed and is pointing to an angle of $180^{\circ}$ towards an acoustic source of 4\,kHz. The same principle can be applied to obtain an acoustic image (c,d) when the steering vectors are distributed in 3D space (changing elevation). In (b), the frequency response of a given microphone array can be found. In this case, a sound source is located at an angle of $180^{\circ}$ relative to the microphone array. The frequency response allows the user to identify whether a given microphone array is likely to detect a given set of acoustic frequencies well.

\subsection{Fractional Delays}

The DaS beamforming technique properly delays the input audio samples to generate constructive interference for a particular steered direction. 
These time delays are obtained based on Eq.~\mbox{\ref{eq:discrete_delta_m}} and rounded following Eq.~\mbox{\ref{eq:discrete_delta_m_rounded}}.
The DaS beamforming technique based on these rounded indices provides accurate results when sufficiently high sampling frequencies $F_s$ (i.e. typically beyond 1~MHz) at the DaS stage are chosen. This method, however, may suffer from output degradation in the opposite case ~\cite{maskell1999estimation}, \cite{pedamallu2012microphone}. 
Due to the sampling frequency, the demodulation of the Pulse Density Modulation (PDM) signals and filtering of the PDM Micro Electro-Mechanical Systems (MEMS) microphones there exist an error in the estimation of the time delays corresponding to the phase delays of the microphones signals when applying DaS beamforming.
Variations of this phenomenon has also been observed in early modem synchronization, speech coding, musical instruments modelling and realignment of multiple telecommunication signals~\cite{laakso1996splitting}.  

\begin{figure}
    \centering
    \includegraphics[width=1.0\textwidth]{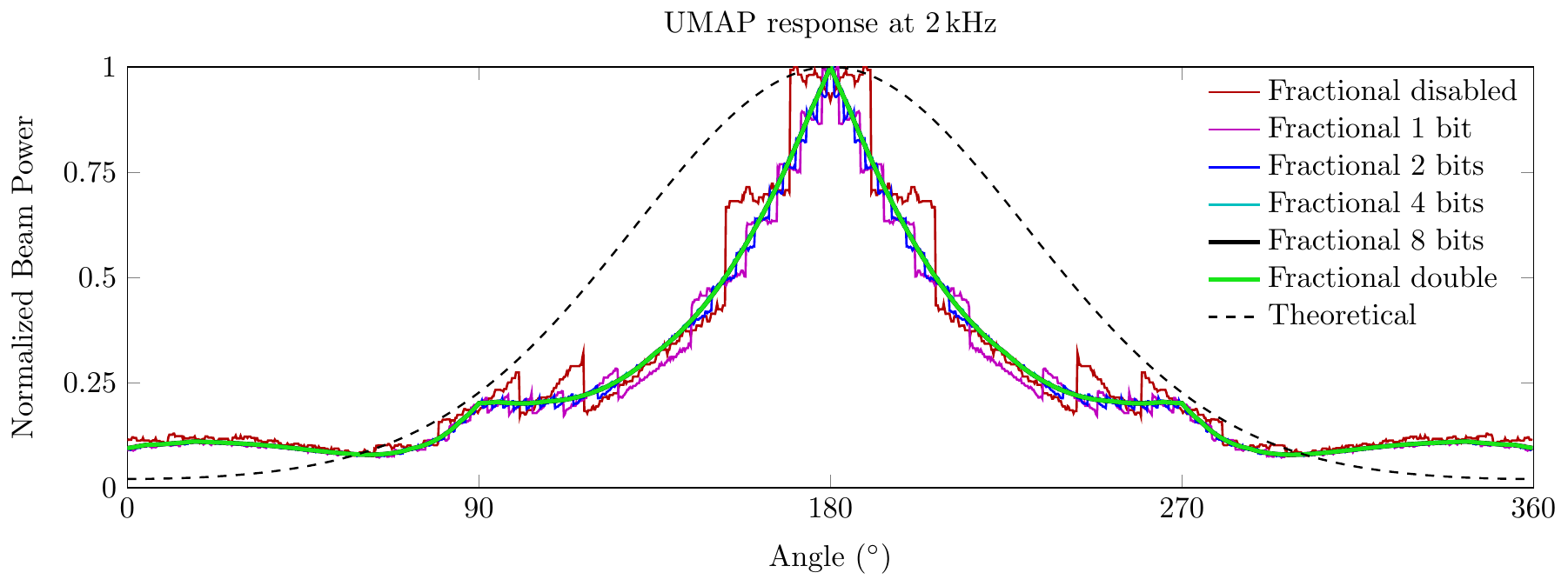}
    \caption{Effects of fractional DaS while detecting a sound source of 2\,kHz with the UMAP microphone array. The theoretical response is given as a reference.}\label{fig:beam_frac_delays_2khz}
\end{figure}

This kind of degradation is shown in Figure \ref{fig:beam_frac_delays_2khz}, where the input sampling frequency is set to 3125\,kHz, but the sampling frequency at the DaS stage is limited to 130,2\,kHz. A sound source is placed at an angle of $180^{\circ}$ from the UMAP microphone array. In the case of the DaS method with index rounding (Fractional disabled, red curve), the microphone array allows the user to find the sound source in an angular area of approximately $30^{\circ}$. However, the staircase like response suggests 2 different closely located sound sources since the graph describes a valley at the supposed $180^{\circ}$ steering orientation. To alleviate this phenomenon, fractional delays can be used to minimize the effects of rounded integer delays and are generally used at the FIR filtering stage~\cite{laakso1996splitting}. Fractional delays can be used in both the time and frequency domains. The method has the advantage of being more flexible in the frequency domain with the added cost of demanding more intense computations. Several time-domain-based implementations do also exist and are generally based on sample interpolation.  

Equations \ref{eq:discrete_delta_m} and \ref{eq:discrete_delta_m_rounded} are rewritten to obtain the floor and the ceiling of the delaying index:  

\begin{equation}
\left\{
\begin{array}{ll}
\Delta'_{m,floor}(\vec{u})&=\left\lfloor F_s\cdot\frac{\vec{r}_m\cdot \vec{u}}{c}\right\rfloor\\[5pt]
\Delta'_{m,ceil}(\vec{u})&=\left\lceil F_s\cdot\frac{\vec{r}_m\cdot \vec{u}}{c}\right\rceil
\end{array}
\right.
\label{eq:discrete_delta_m_floor}
\end{equation}

Based on the floor and the ceiling of the $\Delta'_m$ index, a linear interpolation can be applied at the DaS stage:

\begin{equation}
o[\vec{u},i]_{interp}=\sum\limits_{m=0}^{M-1}\frac{\left(\alpha\cdot s_m[i-\Delta'_{m,floor}(\vec{u})] + (1-\alpha)\cdot s_m[i-\Delta'_{m,ceil}(\vec{u})]\right)}{2} 
\label{eq:sample_double_interpolation}
\end{equation}

With the double precision weighing coefficient $\alpha(\vec{u})$: 

\begin{equation}
\begin{array}{lr}
\alpha(\vec{u})=\Delta'_m(\vec{u})-\Delta'_{m,floor}(\vec{u}), & where\;\alpha\in\mathbb{R}\;and\;[0,1[
\end{array}
\label{eq:alpha_value}
\end{equation}

Double precision computations demand quite some computational power, which is unavailable on constrained devices such as embedded systems, and can lead to intolerable execution times and low frame rates.
Luckily, double precision delays can be changed into fractional delays $\alpha'(n,\vec{u})$. To do so, the double precision weighing coefficient is scaled with the number of bits $n$ used in the fraction and rounded to the nearest natural number. 

\begin{equation}
\begin{array}{lr}
\alpha'(n,\vec{u})=round(\alpha\cdot 2^{n}), & where\;\alpha'(n)\in\mathbb{N}\;and\;[0,2^{n}[
\end{array}
\label{eq:alpha_bitwidth}
\end{equation}

The fractional delays $\alpha'(n)$ range from $zero$ up to $2^{n}-1$. In the case where the rounding function returns $2^{n}$, both the $\Delta'_{m,floor}(\vec{u})$ and the $\Delta'_{m,ceil}(\vec{u})$ are increased with one index, while $\alpha'(n,\vec{u})$ is reset to $zero$. This approach prevents index overlap between 2 rounding areas. The resulting output values $o[\vec{u},i,n]_{interp}$ can be calculated using:

\begin{equation}
o[\vec{u},i,n]_{interp}=\sum\limits_{m=0}^{M-1}\frac{\left(\alpha'(n)\cdot s_m[i-\Delta'_{m,floor}(\vec{u})] + (2^{n}-\alpha'(n))\cdot s_m[i-\Delta'_{m,ceil}(\vec{u})]\right)}{2^{n+1}} 
\label{eq:sample_fractional_interpolation}
\end{equation}

In Equation \ref{eq:sample_fractional_interpolation}, the nominator and denominator are both scaled with a factor $2^{n}$ as compared to Equation \ref{eq:sample_double_interpolation}. The main advantage of the latter is that since the denominator remains a power of 2, a simple bitshift operation can be used instead of a full division mechanism in computationally constrained devices.

The effects of the fractional delays are demonstrated in Figure \ref{fig:beam_frac_delays_2khz}. A higher value of the bitwidth $n$ allows a more fine-grained DaS computation. For values of $n$ beyond 4, the resulting response is almost equal to the response of the DaS with double precision interpolation.

\begin{figure}[t]
\begin{center}
\includegraphics[scale=.65]{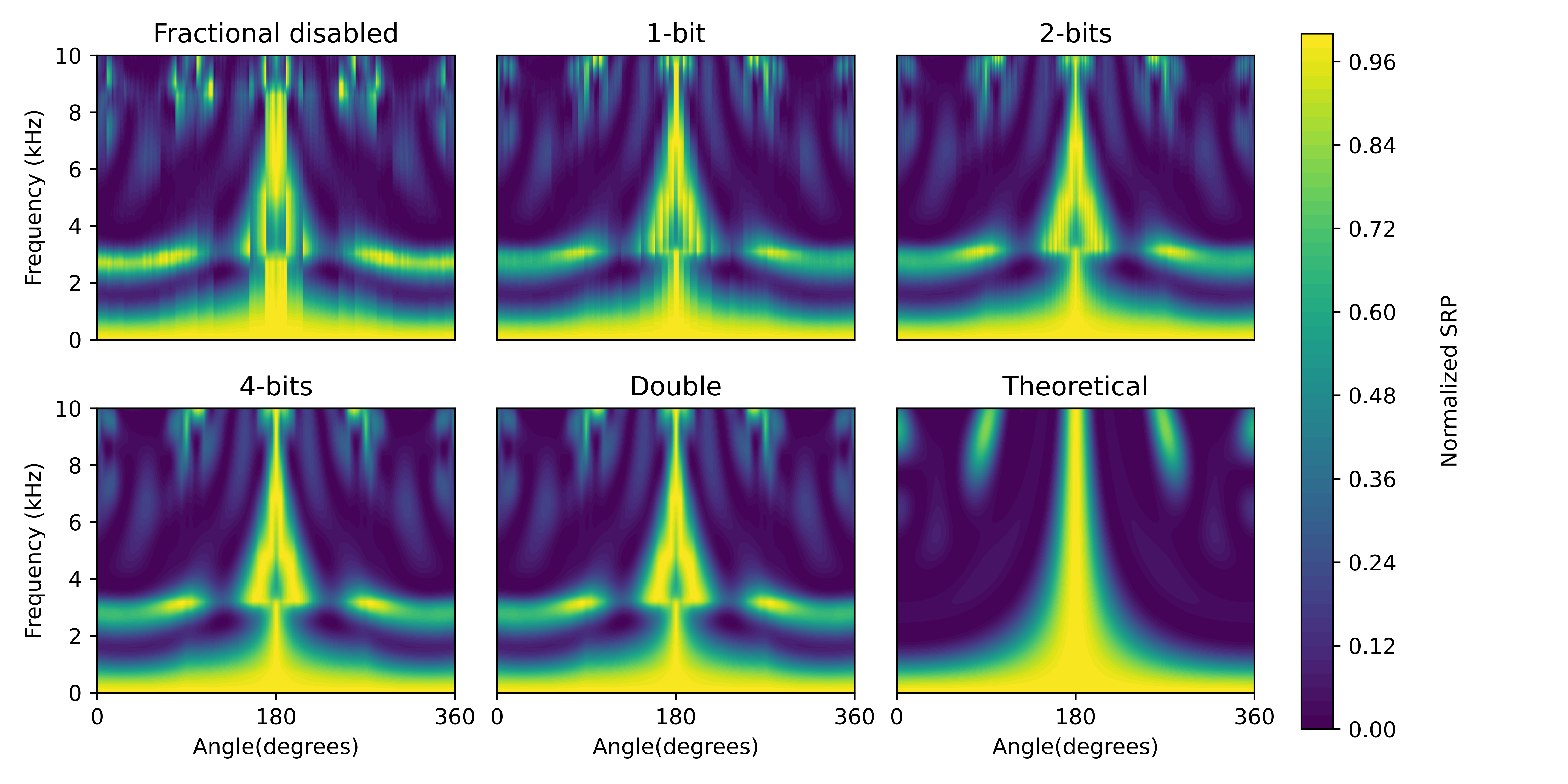}
\end{center}
    \caption{Waterfall diagrams of the UMAP microphone array with different settings of fractional delays, ranging from beamforming without fractional delays (``Fractional disabled'') until fractional delay with a resolution of 8 bits. The methods with 8 bits and double precision fractional delays are represented by the ``double'' since they both give the same results. The waterfall diagram of the theoretical beamforming is given as a reference.}
    \label{fig:waterfall_diagrams}
\end{figure}

In addition to a single response result, the effect of fractional delays is also visible in the waterfall diagrams of the microphone array (Figure \ref{fig:waterfall_diagrams}).
When no fractional delays are used, many vertical stripes are visible, indicating truncation errors during beamforming. By using fractional delays, these stripes gradually disappear until 8 bits (``Fractional 8'') are used. Fractional delays with a resolution of 8 bits and beyond result in the same response as double precision fractional delays. Due to the chosen beamforming architecture, the frequency waterfall diagram of the proposed architecture differs from the theoretically obtainable diagram.  

\section{Acoustic Map Imaging Dataset}
\label{sec:dataset}

This section first introduces the characteristics of the target acoustic camera~\cite{Vandendriessche2021M3AC} and the imaging emulator~\cite{segers2019cabe}. The procedure for capturing multiple scale acoustic map images, the analysis of the dataset, and the applied standardization procedure are also described.

\subsection{Acoustic Camera Characteristics}

\begin{figure}[t]
    \centering
    \includegraphics[width=0.7\textwidth]{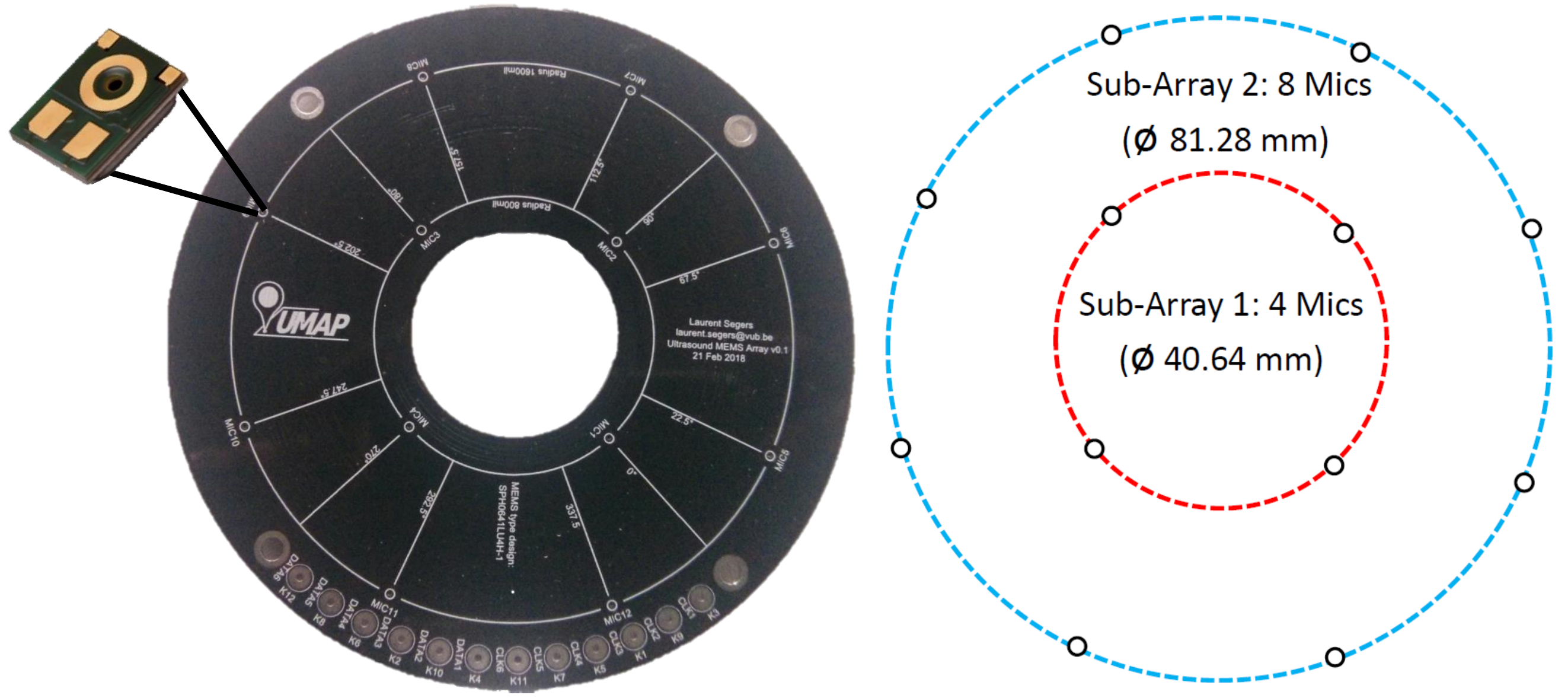}
    \caption{The microphone array used by the M3-AC consists of 12 digital(PDM) microphones. The microphone array is shown in the left image, while the right diagram represents the microphone layout.}
\label{fig:UMAP}
\end{figure}

Acoustic cameras, such as the Multi-Mode Multithread Acoustic Camera (M3-AC) described in~\cite{Vandendriessche2021M3AC}, acquire the acoustic signal using multiple microphones, which convert the acoustic signal into a digital signal. In addition, microphone arrays allow the calculation of the Direction of Arrival (DoA) for a given sound source. The microphone array geometry has a direct impact on the acoustic camera response for DaS beamforming~(Eq.~\ref{eq:steering_all_directions}). 
Figure~\ref{fig:UMAP} depicts the microphone array geometry used by the M3-AC. The microphones are distributed in two circles, with 4 microphones located in the inner circle while the remaining 8 microphones are located in the outer circle. The shortest distance between two microphones is 23.20\,mm and the longest distance is 81.28\,mm.

The type of microphone, the sampling methods, and the signal processing methods influence the final outcome of the beamforming. For instance, the microphone array of M3-AC is composed of MEMS microphones with a PDM output. Despite the benefit of using digital MEMS microphones, there is a need for a PDM demodulation in order to retrieve the acquired audio. Each microphone converts the acoustic into a one-bit PDM signal by using a Sigma-Delta modulator \cite{hegde2010seamlessly}. This modulator typically runs between 1 and 3\,MHz and over-samples the signal. 

\begin{figure}[t]
    \centering
    \includegraphics[width=0.8\textwidth]{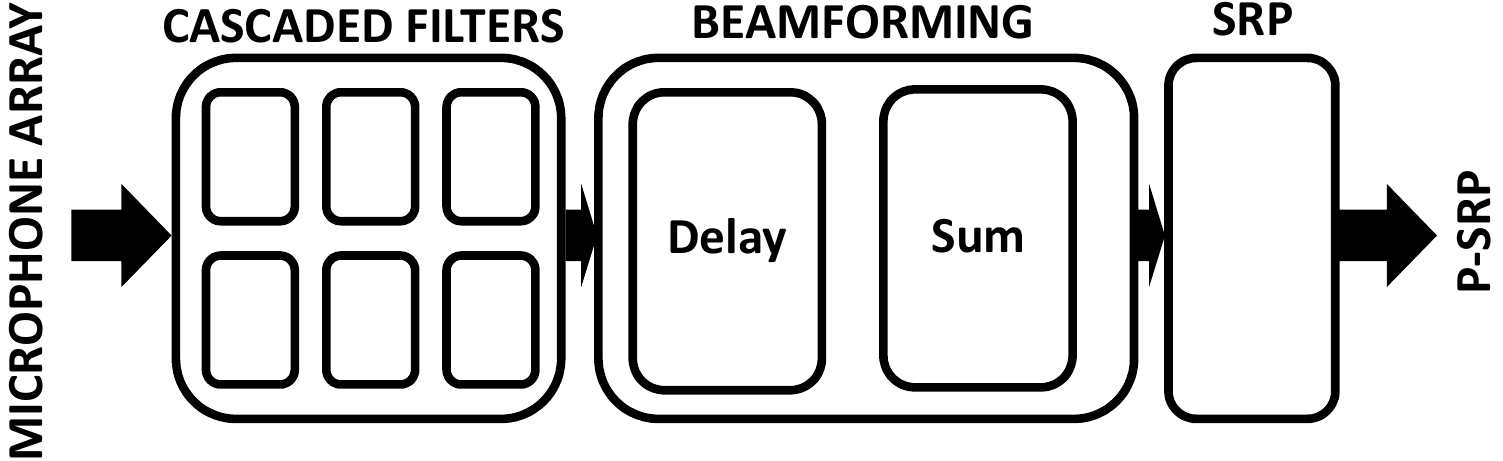}
    \caption{Several cascaded filters are used to demodulate the 1-bit PDM signal generated by the digital MEMS microphones in order to retrieve the original acoustic information in Pulse Coded Modulation (PCM) format. The audio signals are then beamformed by the DaS beamformer. The SRP values are finally used to generate acoustic heatmaps.}
    \label{fig:Architecture}
\end{figure}

To retrieve the encoded acoustic signal, a set of cascaded filters are applied to simultaneously decimate and recondition the signal into a Pulse Coded Modulation (PCM) format (Figure~\ref{fig:Architecture}). Both the geometry of the microphone array and the signal processing of the acoustic signal have a direct impact on the acoustic camera response. 
Evaluation tools such as the Cloud based Acoustic Beamforming Emulator (CABE) enable an early evaluation of an array's geometries and the frequency response of the acoustic cameras ~\cite{segers2019cabe}.

\subsection{Generation of acoustic datasets}

Traditional datasets for SR consist of high-resolution images that are downsampled and blurred to generate the low-resolution image or they use different cameras to generate two images of the same scene~\cite{Rivadeneira_ThermalImageSR}. When using two cameras the frames require some realignment to compensate the different position of the two cameras.

CABE is used to generate the acoustic images.
CABE can emulate the behavior of the traveling sound, microphones and the stages that are required for generating the acoustic heatmap.
The main advantages of using an emulator over real life acoustic heatmaps are consistency and space. First, considering the consistency, using an emulator makes it is possible to replicate the exact same acoustic circumstances multiple times for different resolutions and different configurations.One could generate the same acoustic image with two different microphone arrays for example or use a different filtering stage.

Second, considering space, an emulator eliminates the requirement of having access to anechoic boxes or an anechoic chamber. This allows one to generate acoustic images with sound sources that are several meters away from the microphone array. In order to achieve the same results in a real world scenario, a large anechoic chamber is required. If one has access to such chamber, CABE could still be used. CABE has the option to use PDM signals from real life captured recordings to generate the acoustic heatmaps. Doing so will omit everything that comes before the filtering stage and replace it by the PDM signals from the real life recording.

In order to have a representative dataset, the same architecture as in~\cite{segers2019cabe} is used. The order of the filters and the decimation factor can be found in Table~\ref{tbl:CABE_Settings}.
To compensate for the traveling time of the wave, each emulation starts after 50~ms and lasts for 50~ms. For all emulations a Field of View (FOV) of 60\degree{} in both directions is used.

\begin{table}
    \centering
    \begin{tabular}{|l|l|}
        \hline
        \textbf{Parameter}    & \textbf{Value}                      \\
        \hline
        Microphone array      & UMAP                                \\
        Beamforming method    & Filtering + Delay and Sum           \\
        Filtering method      & 3125khz\_cic24\_fir1\_ds4           \\
        \hline
        Sampling Frequency ($F_S$)  & 3125kHz   \\
        Order CIC Filter ($N_{CIC}$) & 4 \\
        Decimation factor CIC Filter ($D_{CIC}$) & 24 \\
        Order FIR Filter ($N_{FIR}$) & 23 \\
        Decimation factor FIR Filter ($D_{FIR}$) & 4 \\
        \hline
        SRP in block mode     & yes                                 \\
        SRP length            & 64                                  \\
        Emulation start  time & 50 ms                               \\
        Emulation end time    & 100 ms                              \\
        \hline
    \end{tabular}
    \caption{Configuration used in CABE to generate the acoustic images.}
    \label{tbl:CABE_Settings}
\end{table}

\subsection{Dataset Properties}

\begin{figure}[t]
\begin{center}
    \resizebox{1.1\textwidth}{!}{\input{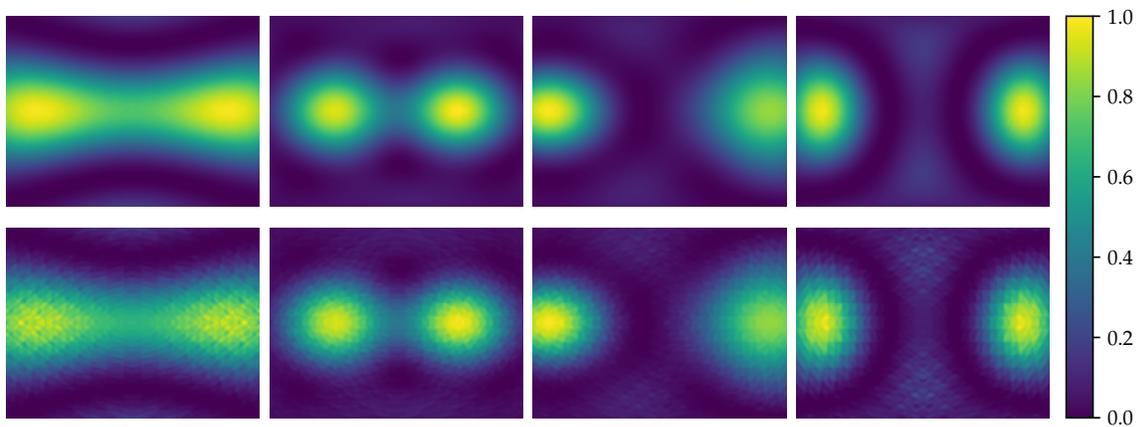}}
\end{center}
  \caption{Acoustic map examples of the test set with double precision high-resolution (top) and fractional delays precision high-resolution (bottom).}
\label{fig:dpvsndp}
\end{figure}

\textbf{Scale}: 
The key application of our proposed acoustic map imaging dataset is to upscale the spatial resolution of a low-resolution image by multiple scale factors (x2, x4, x8). To realise this, 8 different sets of images were generated, each containing 4624 images. Four different resolutions were used: the HR ground truth images of size ($640 \times 480$) and three different scale sets of LR images of size ($320 \times 240$, $160 \times 120$ and $80 \times 60$). For each resolution, one dataset was generated using fractional delays and another without fractional delays as shown in Figure~\ref{fig:dpvsndp} with a total of 36992 images.  In real-world use, acoustic sensor resolution is very small and suffer from a sub-sampling error in the phase delay estimation. This results in artifacts and poor image quality compared to the simulated images. The benchmark used was chosen to simulate these real world difficulties in order to enhance the super-resolved images. This is also consistent with the proposed real captured images set. 

\textbf{Real captured images}: Acoustic maps are generated from recordings of the M3-AC acoustic camera\mbox{\cite{Vandendriessche2021M3AC}}.
The sound sources are placed at different angles without any relation to the positions of the sound sources for the acoustic images generated using CABE.
The acoustic heatmaps are generated with the resolutions corresponding to the x2, x4, x8 scale factors without fractional delays and one double precision set representing the corresponding HR ground truth.
Real world captured images could have different natural characteristics.
This poses a problem as models trained on artificial images cannot generalize to another unknown source of image degradation.
The purpose of the real captured images is to evaluate whether the proposed method can generalize over the unknown data.
For this reason, the real captured images are not used during training, instead they are all exclusively used as test data.

\textbf{Acquisition}: 
Each image contains two sound sources, positioned at a distance of one meter from the center of the array and mirrored from each other.
The sound sources are placed at angles between 60\degree{} and 90\degree{} in steps of 2\degree{} for a total of 16 positions.
No vertical elevation was used.
The frequency of the two sound sources are changed independently of each other from 2~kHz to 10~kHz in steps of 500~Hz, across the 8 different sets.
By using two sound sources, some acoustic images suffer from problems with the angular resolution (Rayleigh criterion) where the distance between the two sound sources becomes too small to distinguish one from another.
For instance, when both sound sources are placed at 90\degree{}, they overlap and become one sound source.

\textbf{Normalization}: 
Natural images in SISR problem are logged in uint8 data type, which is in agreement with the most recent thermal image benchmarks. Although acoustic and thermal sensors allow generation of raw data logged in float representation, which consists of rich information, this could produce a problem in the validation consistency between benchmarks. To avoid technical problems with the validation, the proposed benchmark was standardized with the current research line to be compatible with published datasets in other SISR domains.

The registered sound amplitude in acoustic map imaging depends on the sound volume, the chosen precision during the computation, and the number of microphones. The more microphones that are used during the computations, the higher the amplitude will be in the registered map. This can generate high variant values in the minimum-maximum value range. Due to this, it is not possible to preserve the local dynamic range of the images and to normalize them using fixed global values. It may also cause a problem with the unknown examples.  Any method of contrast or histogram equalization could harm the images and cause loss of important information. Consequently, instance min-max normalization as in Eq. \ref{eq:1} and uint8 data type were used. We denote $I$ and $\bar{I}$ the original image and its normalized version, $\bar{I}_{max} = 1$ and $\bar{I}_{min} = 0$. The images are then converted to grayscale in the range [0 - 255] and saved as PNG format with zero compression.

\begin{equation}
    \bar{I} = (I - I_{min}) \frac{\bar{I}_{max} - \bar{I}_{min}}{I_{max} - I_{min}}   + \bar{I}_{min}
  \label{eq:1}
\end{equation}

\begin{figure}[t]
\begin{center}
  \includegraphics[width=1\linewidth]{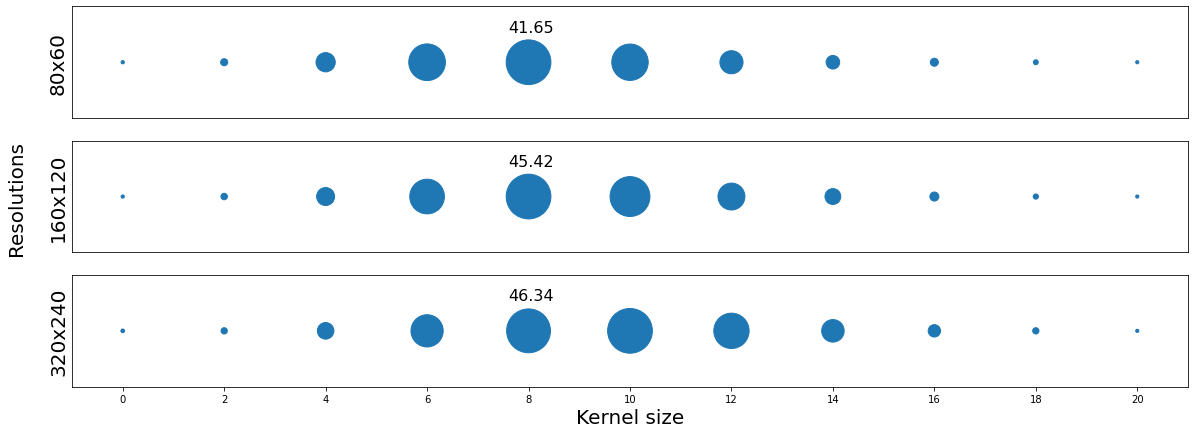}
\end{center}
  \caption{PSNR distribution of the upscaled test set images without fractional delays with scales factor of (x2, x4, x8). The upscaling is done with the bicubic interpolation operator. Gaussian kernels ranging from $0$ to $20$ are used on top to smooth the results. The upscaled images with Gaussian kernel of size 8 achieved the best PSNR results for the three resolutions.}
\label{fig:psnr}
\end{figure}

\textbf{Baseline approach}: 
The authors believe that this is the first work that provides as large as a dataset of acoustic map imaging pairs captured with four different scales in an emulator and in the real-world. Peak signal-to-noise ratio (PSNR) and the structural similarity index measure (SSIM) metrics are reported for reference in SISR problems. A bicubic algorithm is used as a baseline model for validation comparison for the super-resolved images. To reduce quality degradation caused by subsampling errors in phase delay estimation, a Gaussian Kernel with different kernel sizes was used on top of the Bicubic output to reduce the artifacts. Figure.~\ref{fig:psnr} shows that a Gaussian kernel with size 8 achieved the best PSNR results for the three resolutions.

\textbf{Train and Test set}:
Because of the large number of images in the dataset, and to avoid possible overfitting due to shared similarities in the images, 96 samples were drawn for the test set from the images with low PSNR value. The test set sampling procedure is processed on the PSNR distribution built on the 8x bicubic upscaled images. The sampling toward the low PSNR value images is processed using a Kernel Density Estimator (KDE) skewed distribution. In the end, the test set became biased with more complex examples.


\section{XCycles Backprojection Network (XCBP)}
\label{sec:NN}

\subsection{Network Architecture}

\begin{figure}[t]
\begin{center}
  \includegraphics[width=1\linewidth]{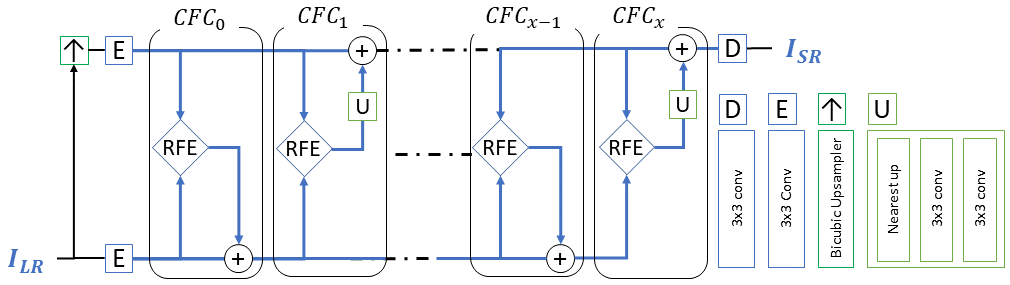}
\end{center}
  \caption{XCycles Backprojection Network architecture.}
\label{fig:model}
\end{figure}

The baseline model of the proposed method comprises two main modules: Cycle Features Correction (CFC) and Residual Features Extraction (RFE). As shown in Figure~\ref{fig:model}, the architecture of XCBP is unfolded with $X$ number of CFCs, as each cycle contains one RFE module.The value of $x$ is an odd number since two consecutive cycles are mandatory for the final results. The model uses only one convolutional layer (Encoder/E) to extract the shallow features $F$ from the low-resolution input image $I_{LR}$ and its pre-upsampled version $ \uparrow I_{LR}$, shown in Eq. \ref{eq:fex}. The pre-usampling module can take any method, such as a classical pre-defined upsampling operator, transposed convolution \cite{dong2016accelerating}, sub-pixel convolution \cite{shi2016real} or resize-convolution \cite{dumoulin2016learned}.

\begin{equation}
\begin{aligned}
  &F_{LR,0} = E(I_{LR}) \\
  &F_{SR,0} = E(\uparrow I_{LR})
\label{eq:fex}
\end{aligned}
\end{equation}

The term $F_{LR,0}$ denotes the encoded features in the low-resolution space, whereas $F_{SR,0}$ denotes the encoded features of the pre-upsampled image in the high-resolution space. The (Decoder/$D$) of only one convolutional layer uses the final features $F_{SR,X}$ corrected by the CFC in cycle $X$ to reconstruct the super-resolved image $I_{SR}$. 

\begin{equation}
\begin{aligned}
  &I_{SR} = D(F_{SR,X})
\end{aligned}
\end{equation}

The CFC module serves as a feature correction mechanism. It is designed to supply the encoded features of the two parallel features spaces to its RFE module for further feature extraction, and corrects the encoded features once at a time. The output of its RFE module is backprojected by addition to one of the two parallel features spaces. The backprojection serves to correct the previous features location in both encoded features manifold in contrast to \cite{rivadeneira2020thermal}. By having the two features spaces as input, the model uses the $F_{LR,x}$ encoded features and its corresponding super-resolved features $F_{SR,x}$ to find the best correction in each feature space. This correction is very helpful for images captured with different devices, with different geometrical registration, that suffer from a misalignment problem. 

The CFC adds the output in an alternate cycle. For each cycle it adds the correction either to the low-resolution features space $F_{LR,x}$ or to the super-resolved features space $F_{SR,x}$. In the $F_{SR,x}$, the output of the RFE passes by the (Upsampler/$U$) before adding the correction to match the scale of the features, as shown in Eq. \ref{eq:alternatives}.
\begin{equation}
\begin{aligned}
  &F_{LR,x} = F_{LR,x-1} + CFC(F_{LR,x-1}, F_{SR,x-1}) \\
  &F_{SR,x} = F_{SR,x-1} + U(CFC(F_{LR,x-1}, F_{SR,x-1}))
\label{eq:alternatives}
\end{aligned}
\end{equation}

The (Upsampler/$U$) is a resize-convolution ~\cite{dumoulin2016learned} sub-module consisting of a pre-defined nearest-neighbor interpolation operator of scale factor $x2$, and a convolution layer with a receptive field of size $5x5$ pixels represented by two stacked $3x3$ convolutions.

\subsection{Residual Features Extraction Module}

The Residual Features Extraction module in each CFC depicted as (RFE) in Figure~\ref{fig:model} is designed to extract features from the two parallel features spaces $F_{LR,x}$ and $F_{SR,x}$. After each cycle correction in one of the features spaces, the encoded features change their characteristics and allocate a new location in the feature space. The RFE module takes both features as input and extracts new residual features for the next feature space correction procedure, based on similarity and non-similarity between previously corrected features spaces.

\begin{figure}[t]
\begin{center}
  \includegraphics[width=1\linewidth]{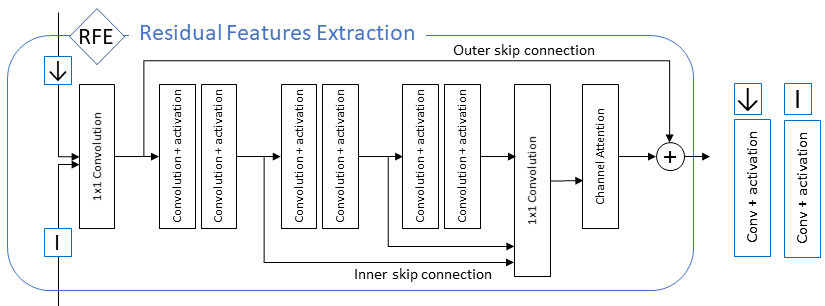}
\end{center}
  \caption{Residual Features Extraction Module}
\label{fig:module}
\end{figure}

As depicted in Figure~\ref{fig:module}, the RFE module has two identical sub-modules (internal features encoder/$I$) responsible for extracting deep features from the two parallel spaces. One ($I$) for each of the features spaces of only one convolutional layer, with straided convolution in the high-resolution space to adapt to the different resolution scale. The two deep encoded features are then concatenated, and a pointwise convolution layer ~\cite{lin2013network} transforms them to their original channel space size.  

The main core of the RFE module consists of three levels $L$ of double activated convolution layers connected sequentially. The output of the three $L$ levels, defined as inner skip connections, are concatenated together, creating dense residual features before a pointwise convolution layer returns them to their original channel space size. Finally, the output of the merger layer is fed to a channel attention module, inspired by the RCAN model, ~\cite{zhang2018image} to weight each residual channel priority before it is added to the outer skip connection of the main merger. 

\subsection{Implementation Details}

The final XCBP proposed model is constructed with X = 8 cycles. All convolution layers are set to 3x3 except for the channel reduction, whose kernel size is 1x1, to transform the concatenated features to their original channel space size. All features are encoded and decoded with 128 feature channels. Convolution layers with kernel size 3x3 uses reflection-padding strategy and Prelu \mbox{\cite{he2015delving}} activation function when activation is stated. The reduction ratio is set to 16 in the channel attention module.

\section{Experiments}
\label{experiments}

\subsection{Training settings}

The experiments are implemented in Pytorch 1.3.1 and performed on an NVIDIA TITAN XP. Ninety percent of the training images were selected, with a total of 4,067 image pairs. Data augmentation is performed on the training images with random rotation of $90^{\circ{}}$, horizontal and vertical flip. A single configuration was used for all experiments and all scale factors. The Adam optimizer and L1 loss were adopted  \cite{kingma2014adam} using default parameter values of zero weight decay, and a learning rate initialized to $10^{−4}$ with step decay of $\gamma = 0.5$ after 500 epochs. The output SR image size for all experiments is $192 \times 192$ with a minibatch size of 8 batches. 

After every epoch the validation metric is run on the current model, and the model with the highest PSNR value is recorded for inference. The model of scale x2 is trained first. Subsequently, the model is frozen and the model of scale x4 is added and trained. The same procedure goes for the model of scale x8.

\subsection{Results}

Table \ref{tbl:psnr_results} shows a quantitative comparison of the best average score for the $\times$2, $\times$4, and $\times$8 super-resolved images compared to the baseline methods: bicubic (MATLAB bicubic operator is used in all experiments), bicubic with a Gaussian of kernel eight, and deep learning SoTA models. The proposed model outperformed the baselines in all of the experiments with significant results. It is also important to note that the model achieved better results on the real captured images as compared with the baselines. This demonstrated that the proposed method could generalize on an unknown acoustic map imaging distribution.

    \begin{table}
    \centering
    \begin{tabular}{|c|c|c|c|c|c|c|}
    \hline
    \multirow{2}{*}{methods} & \multicolumn{3}{c|}{simulated} & \multicolumn{3}{c|}{real captured} \\ \cline{2-7} 
                             & scale x2 & scale x4 & scale x8 & scale x2   & scale x4  & scale x8  \\ \hline
    bicubic &  38.00/0.9426 &  38.16/0.9548 &  37.81/0.9728 &  37.36/0.9513 & 37.31/0.9615 &  36.93/0.9764 \\ \hline
    bicubic-gaussin & 46.34/0.9942 & 45.47/0.9943 & 41.48/0.9935 & 40.99/0.9954 & 40.46/0.9954 & 38.82/0.9946 \\ \hline
    
    SRCNN & 47.00/0.9934 & 46.49/0.9941 & 44.87/0.9938 &
    42.24/0.9940 & 42.25/0.9941 & 42.07/0.9943 \\ \hline
    
    VDSR  & 50.98/0.9963 & 50.89/0.9963 & 49.98/0.9954 &
    44.23/0.9952 & 44.28/0.9950 & 43.470.9942 \\ \hline
    
    
    RCAN & 54.65/0.9978 & 55.19/0.9980 & 54.63/0.9978 &
    \textbf{46.82/0.9971} & 46.57/0.9967 & \textbf{48.88/0.9962} \\ \hline
    
    XCBP-AC & \textbf{54.83/0.9977} & \textbf{55.49/0.9979} &
    \textbf{55.77/0.9980} & 44.64/0.9970 & \textbf{46.66/0.9970} &  46.58/0.9968 \\ \hline
    \end{tabular}
    \caption{Average PSNR/SSIM comparison for x2, x4 and x8 scale factors in the test set between our solutions,  interpolation operators and STOA methods. Best numbers are shown in bold.}
    \label{tbl:psnr_results}
\end{table}



\begin{figure}
\begin{center}
  \includegraphics[width=0.92\linewidth]{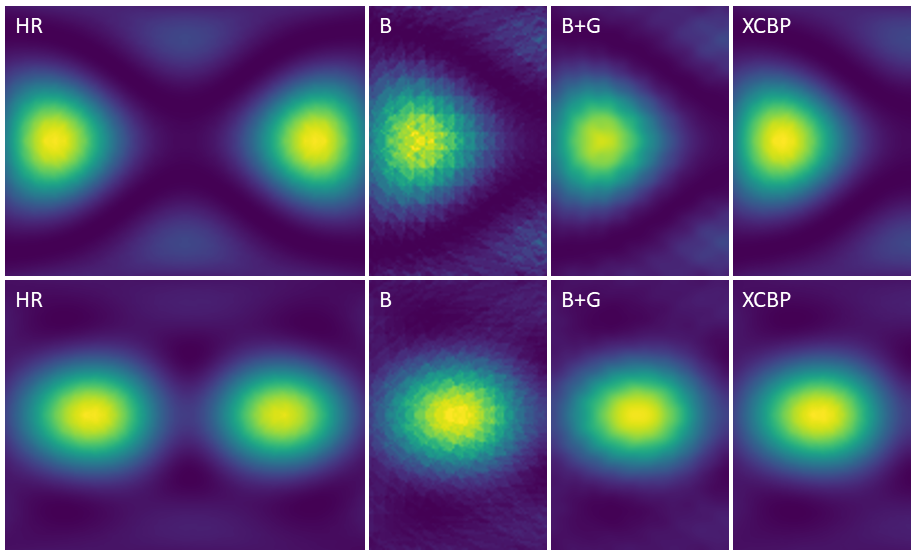}
\end{center}
  \caption{Visual comparison on a cropped region of x4 SR results between: (HR) ground truth high-resolution image. (B) bicubic upscaled image. (B+G) Bicubic and Gaussian upscaled image. (XCBP) our model upscaled image.}
\label{fig:resultsX4}
\end{figure}

\begin{figure}
\begin{center}
  \includegraphics[width=0.92\linewidth]{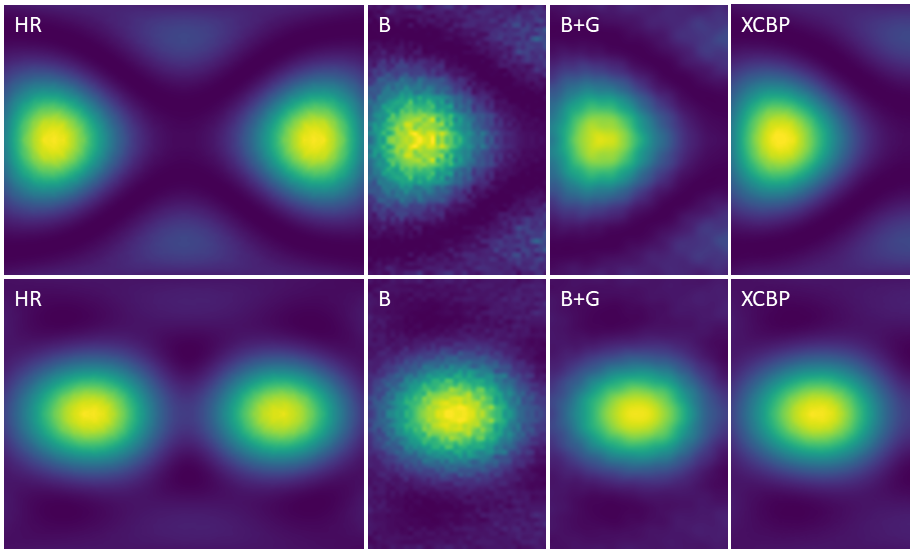}
\end{center}
  \caption{Visual comparison on a cropped region of x8 SR results between: (HR) ground truth high-resolution image. (B) bicubic upscaled image. (B+G) Bicubic and Gaussian upscaled image. (XCBP) our model upscaled image.}
\label{fig:resultsX8}
\end{figure}

\begin{figure}
\begin{center}
  \includegraphics[width=1\linewidth]{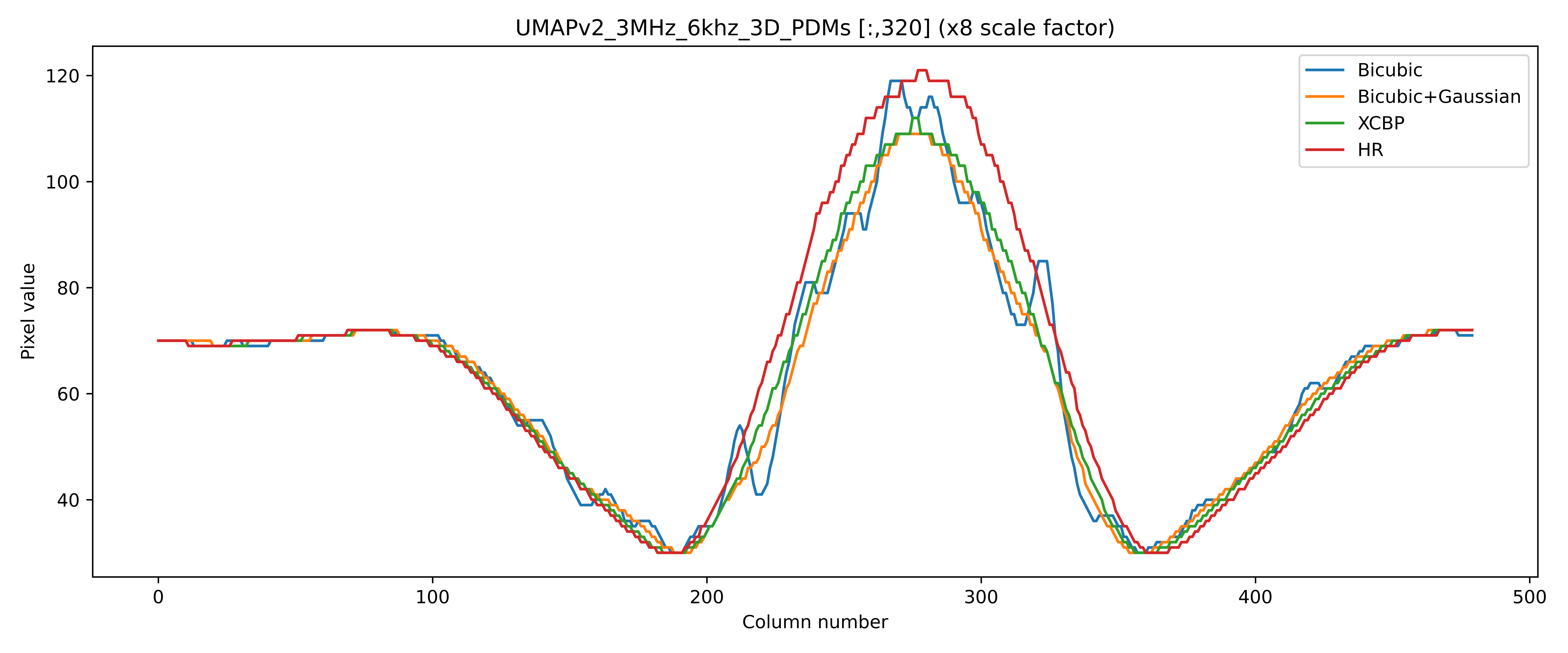}
\end{center}
  \caption{Row profile comparison of real captured x8 SR result. Bucubic operator propagated the sub-sampling error to the SR image creating false positive peak amplitude. Bicubic + Gaussian and XCBP remove the sub-sampling artifacts and look very similar to the HR ground-truth image with XCBP result being the closest output to the HR values.}
\label{fig:rowprofile}
\end{figure}

Despite the quality of the experimental quantitative comparison, the super-resolution problem demands an analytical comparison, as it is not possible to rely only on quantitative standardized metrics such as the PSNR/SSIM. Figures~\ref{fig:resultsX4} and~\ref{fig:resultsX8} show comparisons of the achieved results with the baselines: bicubic upscaled image and bicubic and Gaussian upscaled image on the ($\times$4, $\times$8) scale factors, using images from the simulated test set. It was observed that for the three scaling factors, the proposed model achieved better perceptual results. Given the smooth nature of the acoustic map imaging, applying a Gaussian kernel to the upscaled images greatly enhanced their quality and reduced the artifacts. Although the (bicubic + Gaussian) model achieved excellent results, it was shown that this proposed model surpassed it in the quantitative and analytical comparison.

To confirm further, a row profile test was run on the images to observe the output with the closest result to the ground truth with fewer artifacts. In Figure~\ref{fig:rowprofile}, it is seen that using only the bicubic operator had several false positive sound source maxima because of the sub-sampling artifacts, which were propagated by the operator from the LR image to the super-resolved image, unlike the (bicubic + Gaussian) model and the proposed one, which removed the false positive artifacts and came closer to the ground truth. It was also observed that the proposed model outperformed the (bicubic + Gaussian) model and had higher similarities to the row profile ground truth.

\begin{figure}
    \centering
    \begin{subfigure}[b]{0.23\textwidth}
        \includegraphics[width=\textwidth]{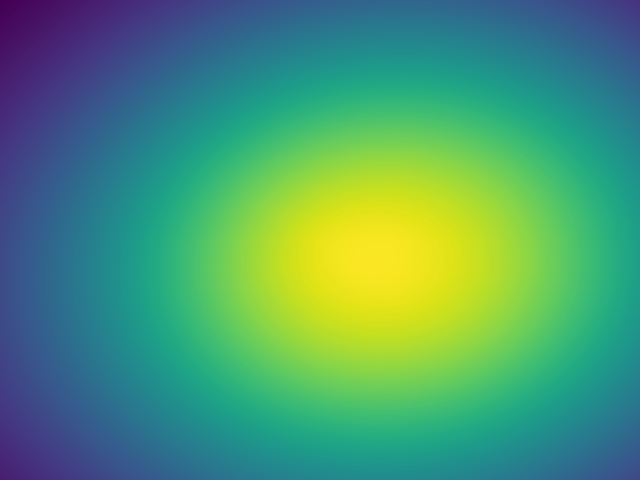}
    \end{subfigure}  
     \begin{subfigure}[b]{0.23\textwidth}
        \includegraphics[width=\textwidth]{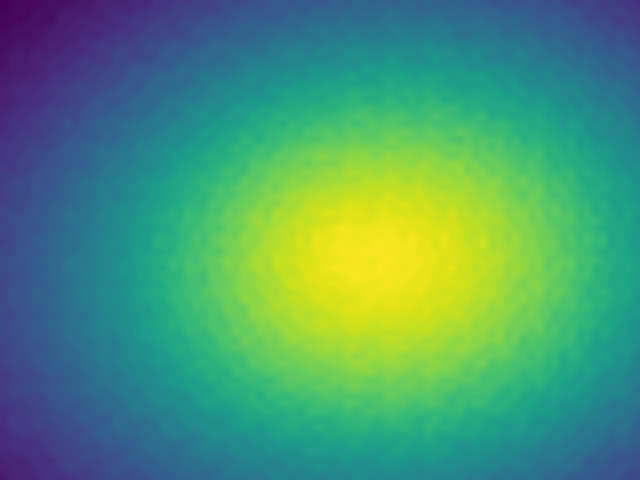}
    \end{subfigure}
    \begin{subfigure}[b]{0.23\textwidth}
        \includegraphics[width=\textwidth]{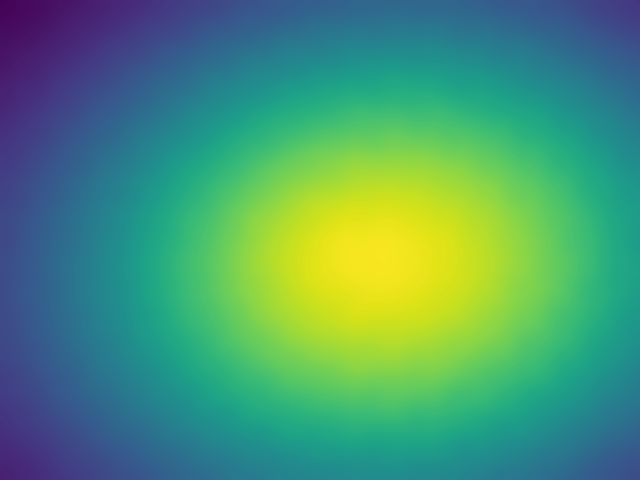}
    \end{subfigure}
    \begin{subfigure}[b]{0.23\textwidth}
        \includegraphics[width=\textwidth]{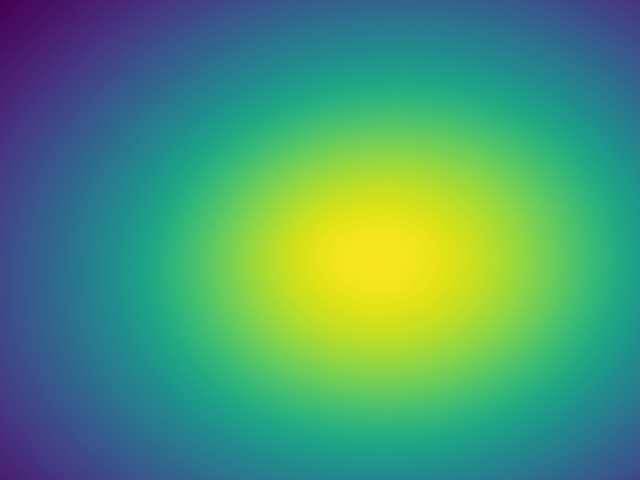}
    \end{subfigure} \\
    \begin{subfigure}[b]{0.23\textwidth}
        \includegraphics[width=\textwidth]{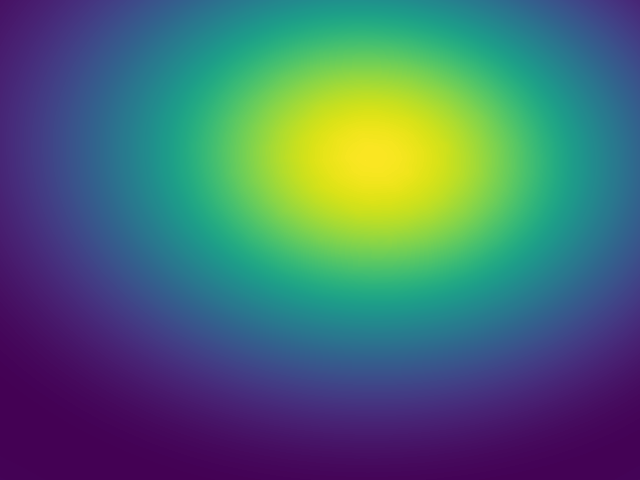}
    \end{subfigure}  
     \begin{subfigure}[b]{0.23\textwidth}
        \includegraphics[width=\textwidth]{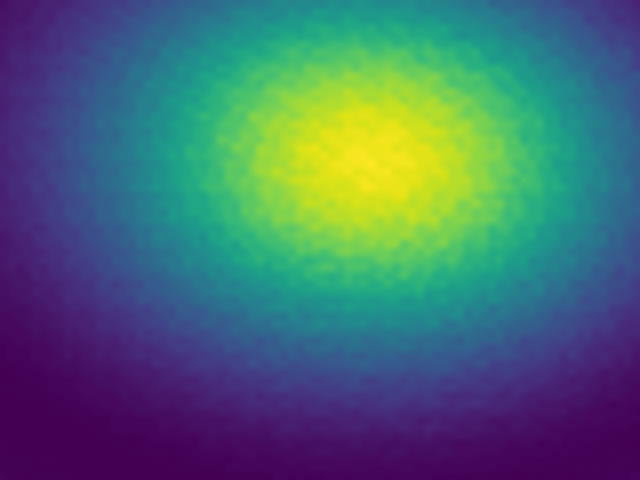}
    \end{subfigure}
    \begin{subfigure}[b]{0.23\textwidth}
        \includegraphics[width=\textwidth]{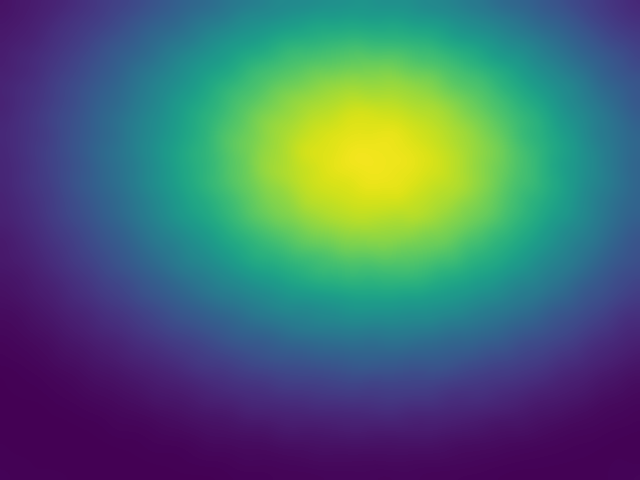}
    \end{subfigure}
    \begin{subfigure}[b]{0.23\textwidth}
        \includegraphics[width=\textwidth]{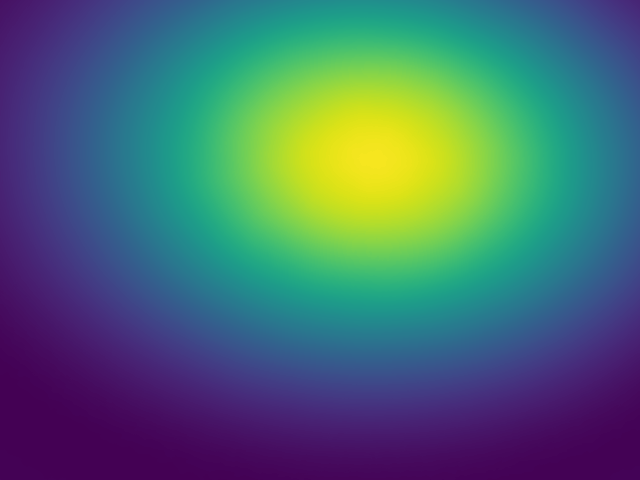}
    \end{subfigure} \\
    \begin{subfigure}[b]{0.23\textwidth}
        \includegraphics[width=\textwidth]{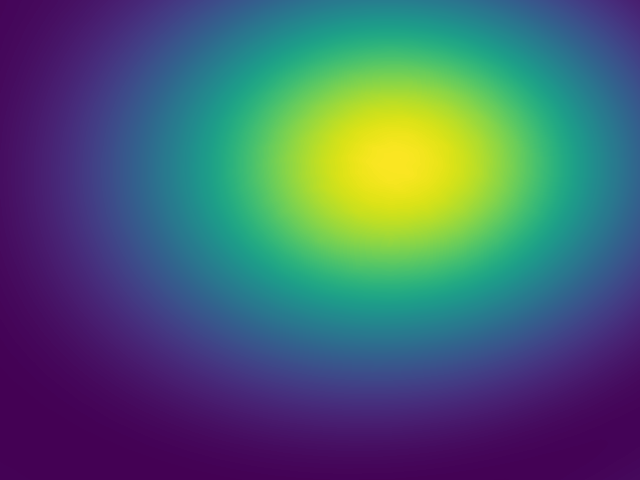}
    \end{subfigure}  
     \begin{subfigure}[b]{0.23\textwidth}
        \includegraphics[width=\textwidth]{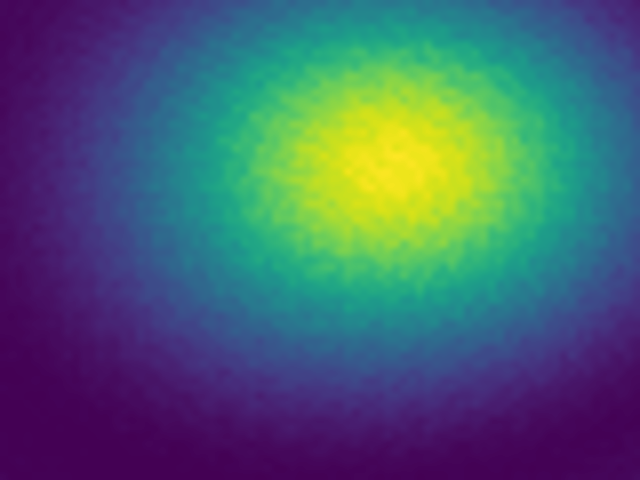}
    \end{subfigure}
    \begin{subfigure}[b]{0.23\textwidth}
        \includegraphics[width=\textwidth]{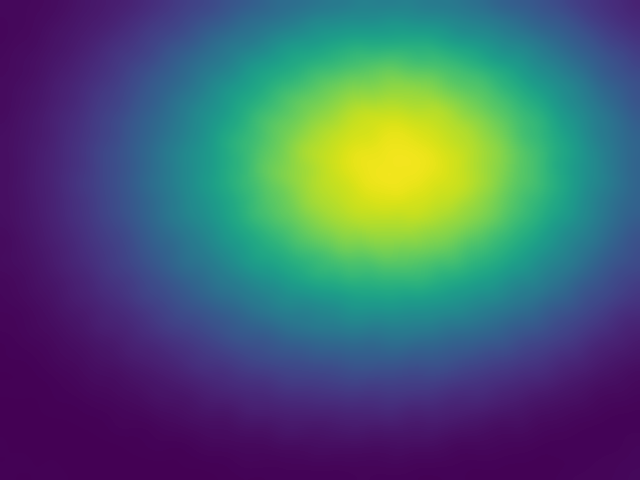}
    \end{subfigure}
    \begin{subfigure}[b]{0.23\textwidth}
        \includegraphics[width=\textwidth]{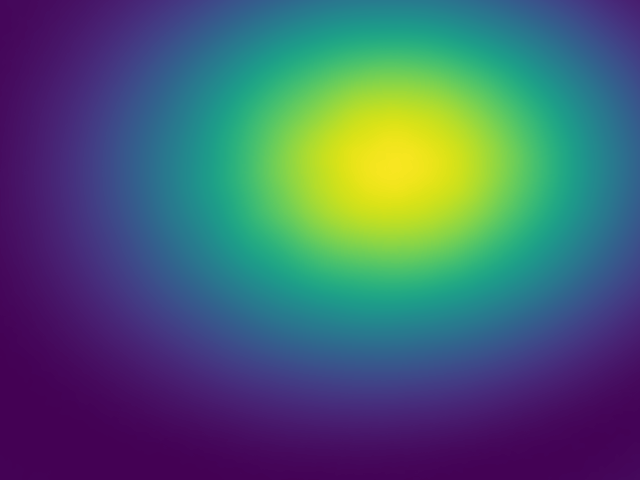}
    \end{subfigure} \\
    \begin{subfigure}[b]{0.23\textwidth}
        \includegraphics[width=\textwidth]{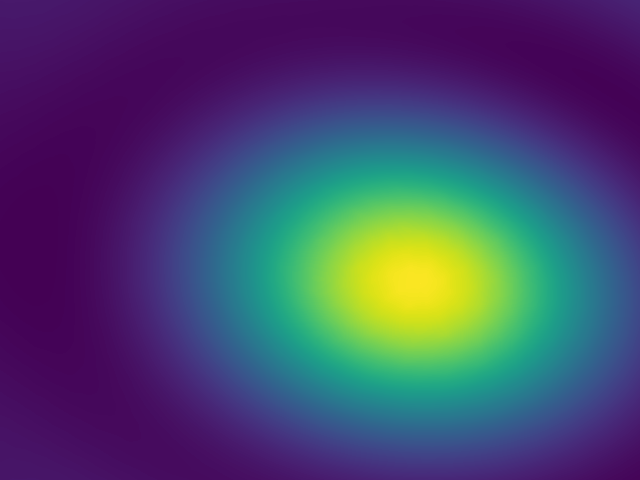}
    \end{subfigure}  
     \begin{subfigure}[b]{0.23\textwidth}
        \includegraphics[width=\textwidth]{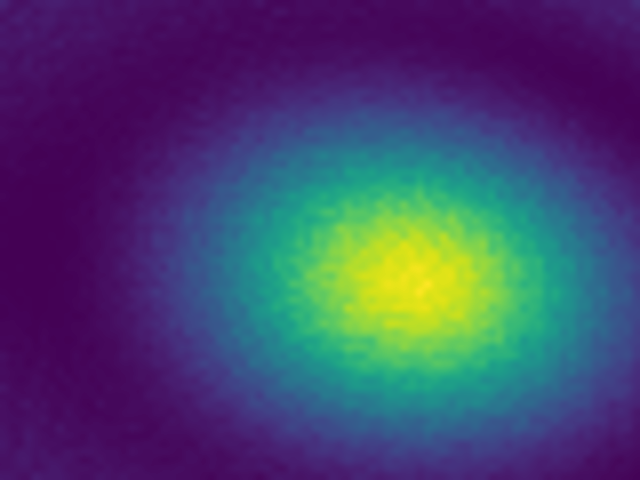}
    \end{subfigure}
    \begin{subfigure}[b]{0.23\textwidth}
        \includegraphics[width=\textwidth]{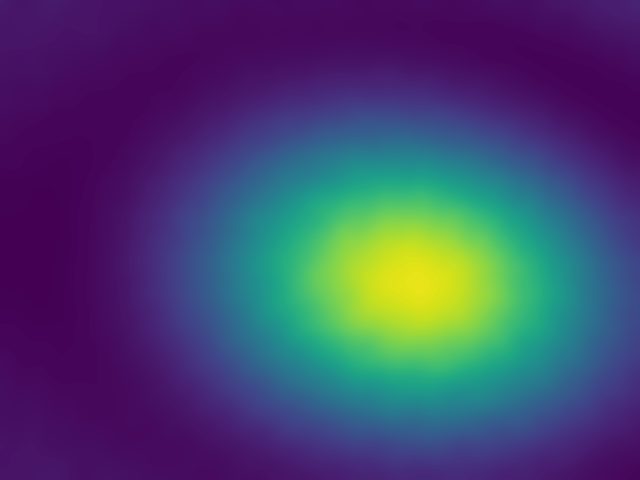}
    \end{subfigure}
    \begin{subfigure}[b]{0.23\textwidth}
        \includegraphics[width=\textwidth]{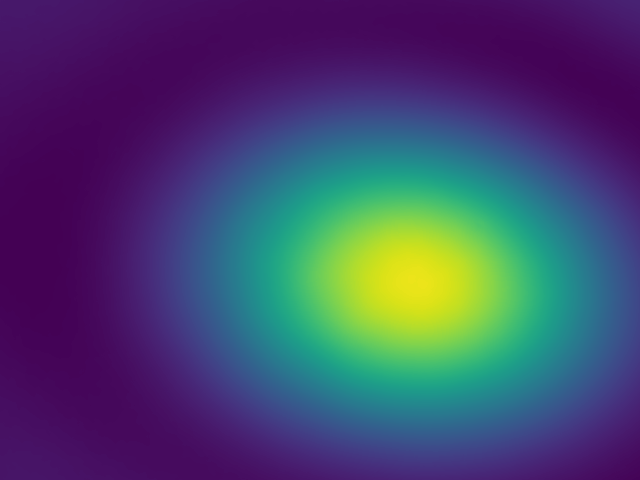}
    \end{subfigure} \\
    \begin{subfigure}[b]{0.23\textwidth}
        \includegraphics[width=\textwidth]{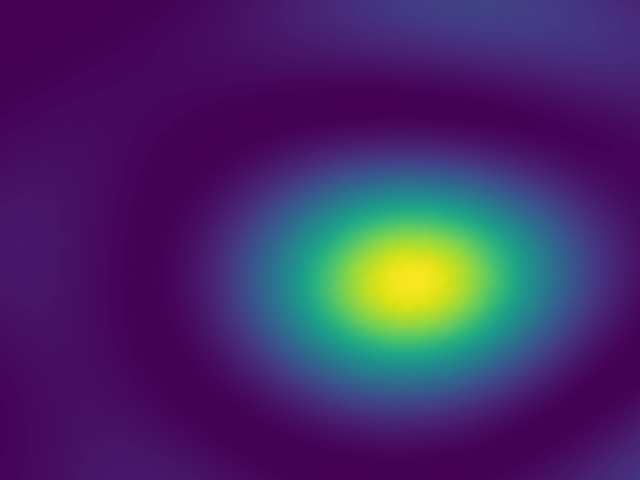}
    \end{subfigure}  
     \begin{subfigure}[b]{0.23\textwidth}
        \includegraphics[width=\textwidth]{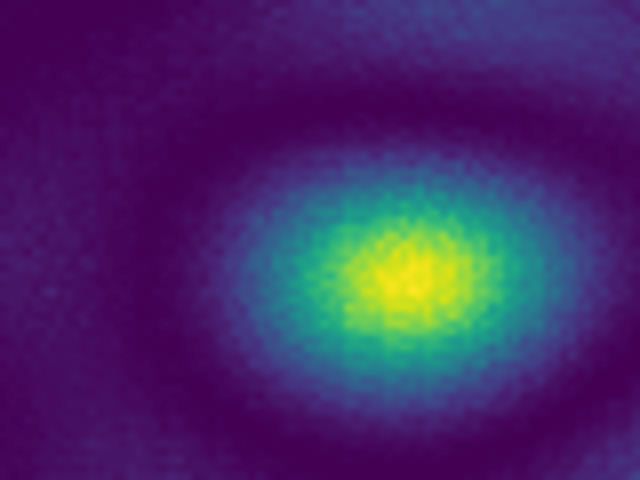}
    \end{subfigure}
    \begin{subfigure}[b]{0.23\textwidth}
        \includegraphics[width=\textwidth]{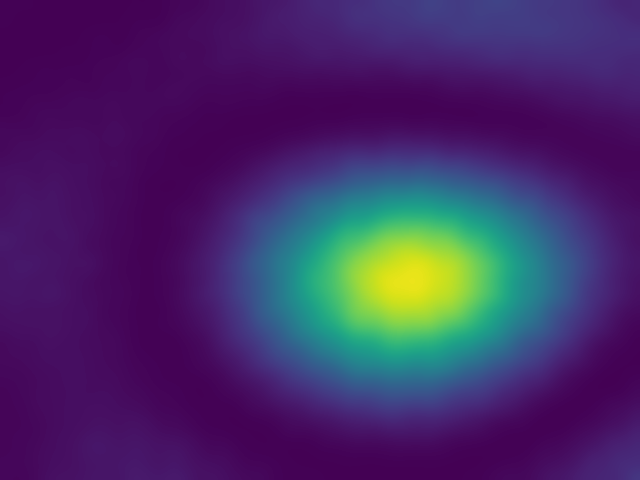}
    \end{subfigure}
    \begin{subfigure}[b]{0.23\textwidth}
        \includegraphics[width=\textwidth]{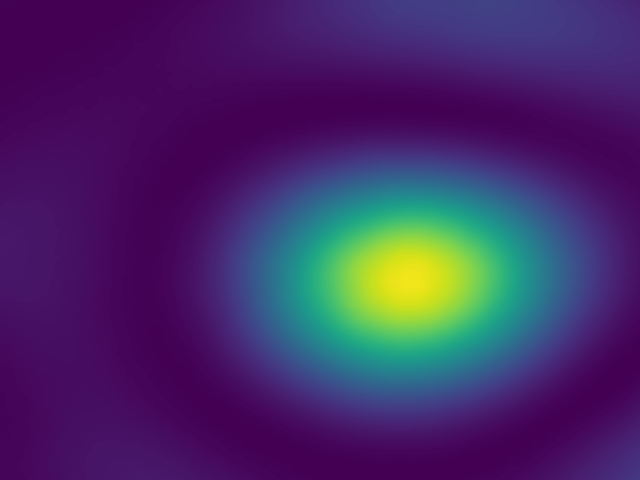}
    \end{subfigure} \\
   
    \vspace{2pt}
 
    \caption{Visual comparison of real captured x8 SR results. Left to right: High-resolution image. Bicubic upscaled image. Bicubic and Gaussian upscaled image. XCBP model upscaled image.}
    \label{fig:realcaptured_x8}
\end{figure}

The obtained images were indeed very similar to the original ground truth HR images. Though this was subjectively confirmed in the cropped regions, the same conclusions can be drawn after observing the entire image such as in Figure~\ref{fig:realcaptured_x8}. Thus, it was shown that this model can upscale the LR image and correct artifacts caused by the sub-sampling error in the three scale factors. 

Both the increase in resolution and the reduction of artifacts helped to improve the quality of the acoustic images and acoustic cameras. First of all, the reduction in noise helped the overall image quality and readability for humans of set images. A second improvement was the frame rate of acoustic cameras. In order to increase the resolution of an acoustic camera without upscaling, one needs to compute more steering vectors and perform more beamforming operations. Beamforming operations are computationally intensive, meaning that increasing the number of steering vectors or the resolution of an acoustic image also increases the computational load and time to generate one acoustic image, giving a trade-off between resolution and frame rate. Using super-resolution to both upscale and reduce the artifacts of nonfractional delays tackled these two problems at the same time. The acoustic camera could generate the acoustic image at a lower resolution and higher frame rate without the need for fractional delays because the super-resolution improved the resolution and quality, without affecting the frame rate.

The super-resolution also allowed better pinpointing the direction sound was coming from and distinguishing sound sources. Outliers caused by the fractional delays were removed, and the increased resolution helped to better estimate the angle of arrival of the sounds.
In order to determine what the origin of the sound was instead of the angle of arrival, acoustic images were combined with images from RGB cameras. RGB cameras have a higher resolution.
To overlay acoustic images with RGB images, both must have the same resolution by increasing the resolution of the acoustic image, decreasing the resolution of the RGB image, or a combination of both. Here, the super resolution can help to match the resolution of the acoustic image with the resolution of the RGB camera.

The similarities in the acquisition of the developed acoustic map imaging dataset may lead to characteristics similarities in the image distribution and to overfitting the data. Given this, the proposed model was tested on real captured images to study its ability to generalize over an unseen data distribution. Although the real captured images still shared characteristics with the simulated images in terms of the number of sound sources and their frequency, they were recorded using different equipment and in another environment, which reduced the possibility of overfitting. Figure~\ref{fig:realcaptured_x8} shows comparisons of the results with the baselines: bicubic upscaled image, bicubic and Gaussian upscaled image on a $\times$8 scale factor, and using real captured images from the test set. It was observed that the proposed model was more capable of upscaling images and reducing artifacts on unseen data as compared to other interpolation operators.

\section{Conclusions}
\label{sec:conclusions}
This work proposed the XCycles BackProjection model (XCBP) for highly accurate acoustic map image super-resolution. The model extracts the necessary residual features in the parallel (HR and LR) spaces and applies feature correction. The model outperforms the current state-of-the-art models and drastically reduces the sub-sampling phase delay error estimation in the acoustic map imaging. An Acoustic map imaging dataset, which provides simulated and real captured images with multiple scale factors (x2, x4, x8) was also proposed. The dataset contains low- and high-resolution images with double precision fractional delays and sub-sampling phase delay error. The proposed dataset can encourage the development of better solutions related to acoustic map imaging.

\vspace{6pt} 


\funding{This work was partially supported by the European Regional Development Fund (ERDF) and the Brussels-Capital Region-Innoviris within the framework of the Operational Programme 2014-2020 through the ERDF-2020 Project ICITYRDI.BRU. This work is also part of the COllective Research NETworking (CORNET) project \textit{"AITIA: Embedded AI Techniques for Industrial Applications"}~\cite{brandalero2020aitia}. The Belgian partners are funded by VLAIO under grant number HBC.2018.0491, while the German partners are funded by the BMWi (Federal Ministry for Economic Affairs and Energy) under IGF-Project Number 249 EBG.}

\dataavailability{The dataset is available online at \url{https://doi.org/10.5281/zenodo.4543785}.}


\reftitle{References}
\externalbibliography{yes}
\bibliography{template}


\end{document}